\documentclass[a4paper]{article}
\usepackage{bbold}
\usepackage{INTERSPEECH2021}
\usepackage{multirow}
\usepackage{xcolor}

\title{Building a great multi-lingual teacher with sparsely-gated mixture of experts for speech recognition}
\name{Kenichi Kumatani$^*$\thanks{$^*$Equal contribution}, Robert Gmyr$^*$, Felipe Cruz Salinas$^*$,\\ Linquan Liu, Wei Zuo, Devang Patel, Eric Sun, Yu Shi}
\address{
  Microsoft}
\email{\{kekumata,rogmyr,ancruzsa\}@microsoft.com}

\begin{document}
\newcommand{\nc}{\newcommand}
\def\dsp{\displaystyle}
\def\lb{\label}
\def\erf#1{(\ref{#1})}
\def\merf#1#2{(\ref{#1}--\ref{#2})}
\def\diag{\mbox{$\text{diag}$}}
\def\real{\mbox{$\text{Re}$}}
\def\imag{\mbox{$\text{Im}$}}
\def\bA{\mbox{$\mathbf{A}$}}

\def\fidx{\mbox{$\omega_m$}}
\def\sIdx{\mbox{$(t)$}}
\def\wIdx{\mbox{$(b)$}}
\def\bXb{\mbox{$\mathbf{X}_b$}}
\def\bXbH{\mbox{$\mathbf{X}_b^H$}}

\def\bAt{\mbox{$\bA^T$}}
\def\bDk{\mbox{$\bSigma_{m}$}}
\def\bDuk{\mbox{$\underline{\bSigma}_{m}$}}
\def\bmutk{\mbox{$\tilde{\bmu}_m$}}
\def\Ast{\mbox{$\bA^{(s)T}$}}

\def\wIdx{\mbox{$(b)$}}
\def\skn{\mbox{$\sigma_{kn}^2$}}
\def\stkns{\mbox{$\tilde{\sigma}_{kn}^{(s)\,2}$}}
\def\mutknssq{\mbox{$\tilde{\mu}_{kn}^{(s)\,2}$}}
\def\muhk{\mbox{$\hat{\mu}_k$}}
\def\bAr{\mbox{$\bA_r$}}
\def\bArt{\mbox{$\bA_r^T$}}
\def\rth{\mbox{$r^{\text{th}}$}}
\def\Arst{\mbox{$\bA_r^{(s)\,T}$}}
\def\tth{\mbox{$t^{\text{th}}$}}
\def\xhts{\mbox{$\hat{x}_t^{(s)}$}}
\def\bal{\mbox{$\mathbf{\alpha}$}}
\def\Ls{\mbox{$\mathcal{L}$}}
\def\Lk{\mbox{$\Lambda_k$}}
\def\Lst{\mbox{$\Ls_t$}}
\def\Lstk{\mbox{$\Ls_{k,t}$}}
\def\Pskt{\mbox{$\Ps_{k,t}$}}

\def\xtn{\mbox{$x_{tn}$}}

\def\Ps{\mbox{$\mathcal{P}$}}
\def\Pst{\mbox{$\Ps_t$}}

\def\xhtns{\mbox{$\hat{x}_{tn}^{(s)}$}}
\def\xht{\mbox{$\hat{x}_t$}}

\def\gij{\mbox{$g_{ij}$}}
\def\anm{\mbox{$a_{nm}$}}
\def\adj{\mbox{$\text{adj}\,$}}
\def\danm{\mbox{$\delta_\alpha[n,m]$}}
\def\hapnm{\mbox{$h_{\alpha\rho}[n,m]$}}
\def\cA{\mbox{$\mathcal{A}$}}
\def\cB{\mbox{$\mathcal{B}$}}
\def\cC{\mbox{$\mathcal{C}$}}
\def\cD{\mbox{$\mathcal{D}$}}
\def\cF{\mbox{$\mathcal{F}$}}
\def\cH{\mbox{$\mathcal{H}$}}
\def\cL{\mbox{$\mathcal{L}$}}
\def\cT{\mbox{$\mathcal{T}$}}
\def\calX{\mbox{$\mathcal{X}$}}
\def\calY{\mbox{$\mathcal{Y}$}}
\def\calZ{\mbox{$\mathcal{Z}$}}
\def\calm{\mathcal{M}}
\def\disp{\displaystyle}

\def\Az{\mbox{$A(z)$}}
\def\Bz{\mbox{$B(z)$}}
\def\Gz{\mbox{$G(z)$}}
\def\rhob{\mbox{$\rho_{b}$}}
\def\thetab{\mbox{$\theta_{b}$}}
\def\Real{\mbox{$\text{Re}\,$}}
\def\Imag{\mbox{$\text{Im}\,$}}

\def\bT{\mbox{$\mathbf{T}$}}
\def\bD{\mbox{$\mathbf{D}$}}
\def\bG{\mbox{$\mathbf{G}$}}
\def\bbG{\mbox{$\mathsf{G}$}}
\def\bbGrls{\mbox{$\bbG_{\text{RLS}}$}}
\def\bGrls{\mbox{$\bG_{\text{RLS}}$}}
\def\bbGk{\mbox{$\bbG_{\text{K}}$}}
\def\bmu{\mbox{$\boldsymbol{\mu}$}}
\def\bvarphi{\mbox{$\boldsymbol{\varphi}$}}
\def\tauh{\mbox{$\hat{\tau}$}}
\def\btheta{\mbox{$\boldsymbol{\theta}$}}
\def\bdelta{\mbox{$\boldsymbol{\delta}$}}
\def\bSigma{\mbox{$\boldsymbol{\Sigma}$}}
\def\bd{\mbox{$\mathbf{d}$}}
\def\taub{\mbox{$\bar{\tau}$}}
\def\epsb{\mbox{$\bar{\epsilon}$}}
\def\btau{\mbox{$\boldsymbol{\tau}$}}
\def\btaub{\mbox{$\bar{\boldsymbol{\tau}}$}}
\def\btauh{\mbox{$\hat{\boldsymbol{\tau}}$}}
\def\bQ{\mbox{$\mathbf{Q}$}}
\def\bbQ{\mbox{$\mathsf{Q}$}}
\def\bbD{\mbox{$\mathsf{D}$}}
\def\bg{\mbox{$\mathbf{g}$}}
\def\blambda{\mbox{$\mathbf{\lambda}$}}
\def\cO{\mbox{$\mathcal{O}$}}
\def\cM{\mbox{$\mathcal{M}$}}
\def\bN{\mbox{$\mathbf{N}$}}
\def\bF{\mbox{$\mathbf{F}$}}
\def\bbF{\mbox{$\mathsf{F}$}}
\def\bZ{\mbox{$\mathbf{Z}$}}
\def\bbZ{\mbox{$\mathsf{Z}$}}
\def\bZfb{\mbox{$\bZ_{\text{fb}}$}}
\def\bz{\mbox{$\mathbf{z}$}}
\def\bb{\mbox{$\mathbf{b}$}}
\def\bbf{\mbox{$\mathbf{f}$}}
\def\bB{\mbox{$\mathbf{B}$}}
\def\bBH{\mbox{$\mathbf{B}^H$}}
\def\bbB{\mbox{$\mathsf{B}$}}
\def\bM{\mbox{$\mathbf{M}$}}
\def\bbm{\mbox{$\mathbf{m}$}}
\def\br{\mbox{$\mathbf{r}$}}
\def\bxh{\mbox{$\hat{\bx}$}}
\def\bX{\mbox{$\mathbf{X}$}}
\def\bbX{\mbox{$\mathsf{X}$}}
\def\bR{\mbox{$\mathbf{R}$}}
\def\bbR{\mbox{$\mathsf{R}$}}
\def\bRx{\mbox{$\bR_{\mathbf{x}}$}}
\def\bY{\mbox{$\mathbf{Y}$}}
\def\bPhi{\mbox{$\mathbf{\Phi}$}}
\def\bzeta{\mbox{$\boldsymbol{\zeta}$}}
\def\bbeta{\mbox{$\boldsymbol{\eta}$}}
\def\etab{\mbox{$\boldsymbol{\eta}$}}
\def\bP{\mbox{$\mathbf{P}$}}
\def\bbP{\mbox{$\mathsf{P}$}}
\def\bPz{\mbox{$\bP_{\mathbf{Z}}$}}
\def\bPc{\mbox{$\bP_{\mathbf{C}}$}}
\def\bPcperp{\mbox{$\bP_{\mathbf{C}}^\perp$}}
\def\bV{\mbox{$\mathbf{V}$}}
\def\bVH{\mbox{$\mathbf{V^H}$}}
\def\bx{\mbox{$\mathbf{x}$}}
\def\by{\mbox{$\mathbf{y}$}}
\def\be{\mbox{$\mathbf{e}$}}
\def\bLambda{\mbox{$\mathbf{\Lambda}$}}
\def\bN{\mbox{$\mathbf{N}$}}
\def\bNH{\mbox{$\mathbf{N^H}$}}
\def\bv{\mbox{$\mathbf{v}$}}
\def\bvk{\mbox{$\mathbf{v}_{\mathbf{k}}$}}
\def\bk{\mbox{$\mathbf{k}$}}
\def\bK{\mbox{$\mathbf{K}$}}
\def\bbK{\mbox{$\mathsf{K}$}}
\def\bS{\mbox{$\mathbf{S}$}}
\def\brhon{\mbox{$\brho_{\mathbf{n}}$}}
\def\brhoni{\mbox{$\brho_{\mathbf{n}}^{-1}$}}
\def\bSQ{\mbox{$\bS_{\mathbf{Q}}$}}
\def\bSc{\mbox{$\bS_{\mathbf{c}}$}}
\def\bSx{\mbox{$\bS_{\mathbf{x}}$}}
\def\bgz{\mbox{$\bg_{\mathbf{Z}}$}}
\def\bSz{\mbox{$\bS_{\mathbf{z}}$}}
\def\bSn{\mbox{$\bS_{\mathbf{n}}$}}
\def\bSf{\mbox{$\bS_{\mathbf{f}}$}}
\def\bSs{\mbox{$\bS_{\mathbf{s}}$}}
\def\bI{\mbox{$\mathbf{I}$}}
\def\bw{\mbox{$\mathbf{w}$}}
\def\bff{\mbox{$\mathbf{f}$}}
\def\bW{\mbox{$\mathbf{W}$}}
\def\bbW{\mbox{$\mathsf{W}$}}
\def\bU{\mbox{$\mathbf{U}$}}
\def\bUH{\mbox{$\mathbf{U}^H$}}
\def\bbU{\mbox{$\mathsf{U}$}}
\def\bXb{\mbox{$\mathbf{X}_b$}}
\def\bXbH{\mbox{$\mathbf{X}_b^H$}}
\def\bu{\mbox{$\mathbf{u}$}}
\def\bp{\mbox{$\mathbf{p}$}}
\def\bpz{\mbox{$\mathbf{p}_{\mathbf{z}}$}}
\def\bpHz{\mbox{$\mathbf{p}^T_{\mathbf{z}}$}}
\def\bc{\mbox{$\mathbf{c}$}}
\def\bss{\mbox{$\mathbf{s}$}}
\def\ba{\mbox{$\mathbf{a}$}}
\def\bH{\mbox{$\mathbf{H}$}}
\def\bh{\mbox{$\mathbf{h}$}}
\def\mth{\mbox{$m$th}}
\def\sinc{\mbox{$\text{sinc}$}}
\def\bphi{\mbox{$\mathbf{\Phi}$}}
\def\brho{\mbox{$\mathbf{\rho}$}}
\def\bzero{\mbox{$\mathbf{0}$}}
\def\mini{\mbox{$\text{minimize}$}}
\def\subj{\mbox{$\text{subject to}$}}
\def\st{\mbox{$\text{s.t.}$}}
\def\nth{\mbox{$n$th}}
\def\kth{\mbox{$k$th}}
\def\lth{\mbox{$l$th}}
\def\dth{\mbox{$d$th}}
\def\ith{\mbox{$i$th}}
\def\jth{\mbox{$j$th}}
\def\Kth{\mbox{$K$th}}
\def\bxtn{\mbox{$\tilde{\bx}_n$}}
\def\bwtn{\mbox{$\tilde{\bw}_n$}}
\def\bWtn{\mbox{$\tilde{\bW}_n$}}
\def\bXtn{\mbox{$\tilde{\bX}_n$}}
\def\bXt{\mbox{$\tilde{\bX}$}}
\def\bl{\mbox{$\mathbf{l}$}}
\def\bL{\mbox{$\mathbf{L}$}}
\def\bXtfb{\mbox{$\bXt_{\text{fb}}$}}
\def\bXtHfb{\mbox{$\bXt^T_{\text{fb}}$}}
\def\bYt{\mbox{$\tilde{\bY}$}}
\def\byt{\mbox{$\tilde{\by}$}}
\def\bFm{\mbox{$\bF_M$}}
\def\bFmi{\mbox{$\bF_M^{-1}$}}
\def\bFmh{\mbox{$\bF_M^T$}}
\def\bwdq{\mbox{$\bar{\bw}_{\text{dq}}$}}
\def\bwdqH{\mbox{$\bar{\bw}_{\text{dq}}^T$}}
\def\balpha{\mbox{$\boldsymbol{\alpha}$}}
\def\bepsilon{\mbox{$\boldsymbol{\epsilon}$}}

\def\tr{\mbox{$\text{tr}$}}
\def\bJ{\mbox{$\mathbf{J}$}}
\def\xtmn{\mbox{$\tilde{x}_m[n]$}}

\def\bOmega{\mbox{$\mathbf{\Omega}$}}

\def\bwnn{\mbox{$\text{BW}_{\text{NN}}$}}
\def\rnn{\mbox{$r_{\text{NN}}$}}

\def\bSxi{\mbox{$\bS_{\mathbf{x}}^{-1}$}}
\def\bSzi{\mbox{$\bS_{\mathbf{z}}^{-1}$}}
\def\bSni{\mbox{$\bS_{\mathbf{n}}^{-1}$}}
\def\bSxt{\mbox{$\bS_{\mathbf{x}}^T$}}
\def\bSxh{\mbox{$\hat{\bS}_{\mathbf{x}}$}}
\def\bC{\mbox{$\mathbf{C}$}}
\def\bxi{\mbox{$\boldsymbol{\xi}$}}
\def\bnu{\mbox{$\boldsymbol{\nu}$}}
\def\bbC{\mbox{$\mathsf{C}$}}
\def\bbCk{\mbox{$\mathsf{C}_{\text{K}}$}}
\def\bbCkH{\mbox{$\mathsf{C}^T_{\text{K}}$}}
\def\bbCrls{\mbox{$\bbC_{\text{RLS}}$}}
\def\bbCrlsH{\mbox{$\bbC^T_{\text{RLS}}$}}
\def\bone{\mbox{$\mathbf{1}$}}
\def\bCx{\mbox{$\bC_{\mathbf{x}}$}}
\def\bCxi{\mbox{$\bC_{\mathbf{x}}^{-1}$}}
\def\bCxt{\mbox{$\bC_{\mathbf{x}}^T$}}
\def\hideal{\mbox{$h_{\text{ideal}}$}}
\def\bSh{\mbox{$\hat{\bS}$}}
\def\bSt{\mbox{$\tilde{\bS}$}}
\def\bStx{\mbox{$\tilde{\bS}_{\mathbf{x}}$}}
\def\bStxi{\mbox{$\tilde{\bS}_{\mathbf{x}}^{-1}$}}
\def\bShn{\mbox{$\hat{\bS}_{\mathbf{n}}$}}
\def\bShx{\mbox{$\hat{\bS}_{\mathbf{x}}$}}
\def\bShz{\mbox{$\hat{\bS}_{\mathbf{z}}$}}
\def\bShni{\mbox{$\hat{\bS}^{-1}_{\mathbf{n}}$}}
\def\bShxi{\mbox{$\hat{\bS}^{-1}_{\mathbf{x}}$}}
\def\bShzi{\mbox{$\hat{\bS}^{-1}_{\mathbf{z}}$}}
\def\bShxfb{\mbox{$\hat{\bS}_{\mathbf{x},\text{fb}}$}}
\def\bShnfb{\mbox{$\hat{\bS}_{\mathbf{n},\text{fb}}$}}
\def\nn{\nonumber}
\def\bl{\mathbf{l}}

\def\qk{\mbox{$w_m$}}
\def\qik{\mbox{$w_{k|i}$}}
\def\zkt{\mbox{$x_{k,t}$}}
\def\zkts{\mbox{$x_{k,t}^{(s)}$}}
\def\zmk{\mbox{$x_{m,k}$}}
\def\zmks{\mbox{$x_{m,k}^{(s)}$}}

\def\cm{\mbox{$c_{m}$}}
\def\cmk{\mbox{$c_{m,k}$}}
\def\cmks{\mbox{$c_{m,k}^{(s)}$}}

\def\Lamk{\mbox{$\Lambda_k$}}
\def\Lami{\mbox{$\Lambda_i$}}
\def\Lamik{\mbox{$\Lambda_{ik}$}}
\def\oneHalf{\mbox{$\frac{1}{2}$}}
\def\oneFourth{\mbox{$\frac{1}{4}$}}
\def\oneSixth{\mbox{$\frac{1}{6}$}}
\def\mukt{\mbox{$\mu_{k}^{T}$}}
\def\mukrstar{\mbox{$\mu^{*}_{kr}$}}
\def\mutkst{\mbox{$\tilde{\mu}_{k}^{(s)T}$}}
\def\muk{\mbox{$\mu_k$}}
\def\Qk{\mbox{$Q_k^{-1}$}}
\def\xt{\mbox{$x_t$}}
\def\X{\mbox{$\mathcal{X}$}}
\def\Y{\mbox{$\mathcal{Y}$}}
\def\Ys{\mbox{$\mathcal{Y}^{(s)}$}}
\def\cY{\mbox{$\mathcal{Y}$}}
\def\cW{\mbox{$\mathcal{W}$}}
\def\cP{\mbox{$\mathcal{P}$}}
\def\cX{\mbox{$\mathcal{X}$}}
\def\cN{\mbox{$\mathcal{N}$}}
\def\cE{\mbox{$\mathcal{E}$}}
\def\cI{\mbox{$\mathcal{I}$}}
\def\cZ{\mbox{$\mathcal{Z}$}}
\def\Z{\mbox{$\mathcal{Z}$}}
\def\ck{\mbox{$c_k$}}
\def\cks{\mbox{$c_m^{(s)}$}}
\def\ckts{\mbox{$c_{m,k}^{(s)}$}}
\def\ckt{\mbox{$c_{k,t}$}}
\def\N{\mbox{$\mathcal{N}$}}
\def\Cm{\mbox{$C_m$}}

\def\muik{\mbox{$\mu_{ik}$}}
\def\mukn{\mbox{$\mu_{kn}$}}
\def\mukl{\mbox{$\mu_{kl}$}}
\def\muhks{\mbox{$\hat{\bmu}_m^{(s)}$}}
\def\muhkns{\mbox{$\hat{\mu}_{mn}^{(s)}$}}
\def\muhkn{\mbox{$\hat{\mu}_{kn}$}}
\def\mut{\mbox{$\tilde{\mu}$}}
\def\mutks{\mbox{$\tilde{\mu}_k^{(s)}$}}
\def\bmutms{\mbox{$\tilde{\boldsymbol{\mu}}_m^{(s)}$}}
\def\bmutm{\mbox{$\tilde{\mathbb{\mu}}_m^{(s)}$}}
\def\muh{\mbox{$\hat{\bmu}$}}
\def\mub{\mbox{$\breve{\bmu}$}}

\def\btwo{\mbox{$\tilde{\bw}_o$}}
\def\btwHo{\mbox{$\tilde{\bw}^T_o$}}
\def\btwa{\mbox{$\tilde{\bw}_a$}}
\def\btwHa{\mbox{$\tilde{\bw}^T_a$}}
\def\bhwa{\mbox{$\hat{\bw}_a$}}
\def\bhwHa{\mbox{$\hat{\bw}^T_a$}}

\def\bwtgsc{\mbox{$\tilde{\bw}_{\text{gsc}}$}}
\def\bwtgscH{\mbox{$\tilde{\bw}_{\text{gsc}}$}}

\def\btk{\mbox{\tt btk}}
\def\numpy{\mbox{\tt NumPy}}
\def\pygsl{\mbox{\tt PyGSL}}
\def\python{\mbox{\tt Python}}

\def\SigX{\mbox{$\Sigma_{\mathbf{X}}$}}
\def\SigY{\mbox{$\Sigma_{\mathbf{Y}}$}}
\def\SigN{\mbox{$\Sigma_{\mathbf{N}}$}}
\def\SigXi{\mbox{$\Sigma^{-1}_{\mathbf{X}}$}}
\def\SigYi{\mbox{$\Sigma^{-1}_{\mathbf{Y}}$}}
\def\SigNi{\mbox{$\Sigma^{-1}_{\mathbf{N}}$}}
\def\SigV{\mbox{$\Sigma_{\mathbf{V}}$}}
\def\SigZ{\mbox{$\Sigma_{\mathbf{Z}}$}}

\def\bSigX{\mbox{$\boldsymbol{\Sigma}_{\mathbf{X}}$}}
\def\bSigXi{\mbox{$\boldsymbol{\Sigma}^{-1}_{\mathbf{X}}$}}
\def\bSigN{\mbox{$\boldsymbol{\Sigma}_{\mathbf{N}}$}}
\def\bSigNi{\mbox{$\boldsymbol{\Sigma}^{-1}_{\mathbf{N}}$}}

% Symbols for MMI Training
\def\xls{\mbox{$\by_{1:K}^{(s)}$}}
\def\xlsT{\mbox{$\by_{1:K}^{(s)T}$}}
\def\nls{\mbox{$g_{1:K}^{(s)}$}}
\def\nts{\mbox{$g_k^{(s)}$}}
\def\wns{\mbox{$w_{1:K_{\text{w}}}^{(s)}$}}
\def\wn{\mbox{$w_{1:K_{\text{w}}}$}}
\def\cs{\mbox{$c_{1:K}^{(s)}$}}
\def\oks{\mbox{$\bo_m^{(s)}$}}
\def\xts{\mbox{$\by_k^{(s)}$}}
\def\xtsT{\mbox{$\bx_k^{(s)T}$}}
\def\sks{\mbox{$\bbs_m^{(s)}$}}
\def\skns{\mbox{$s_{kn}^{(s)}$}}
\def\dks{\mbox{$d_m^{(s)}$}}
\def\muok{\mbox{$\bmu^0_m$}}
\def\mutkns{\mbox{$\hat{\mu}_{mn}^{0(s)}$}}

\def\Dk{\mbox{$\bSigma_{m}^{-1}$}}
\def\Dik{\mbox{$\bSigma_{im}^{-1}$}}

\def\bs{\mbox{$\mathbf{s}$}}

\def\bbA{\mbox{$\underline{\bA}$}}
\def\bbSigma{\mbox{$\underline{\bSigma}$}}
\def\bbbb{\mbox{$\mathbb{b}$}}
\def\bbmutm{\mbox{$\underline{\tilde{\bmu}}_m$}}
\def\As{\mbox{$\bA^{(s)}$}}
\def\bAs{\mbox{$\bA^{(s)}$}}
\def\Ars{\mbox{$\bA_r^{(s)}$}}
\def\Aos{\mbox{$\bA_0^{(s)}$}}
\def\AsT{\mbox{$\bA^{(s)\,T}$}}
\def\ArsT{\mbox{$\bA_r^{(s)\,T}$}}
\def\cSOne{\mbox{$S^{(1)}$}}
\def\cSTwo{\mbox{$S^{(2)}$}}
\def\brs{\mbox{$b_r^{(s)}$}}
\def\bos{\mbox{$b_0^{(s)}$}}
\def\Dki{\mbox{$D_k^{-1}$}}
\def\muhko{\mbox{$\hat{\mu}_k^0$}}
\def\miT{\mbox{$m_i^T$}}
\def\mjT{\mbox{$m_j^T$}}
\def\fiT{\mbox{$f_i^T$}}
\def\ots{\mbox{$\bo_{m,k}^{(s)}$}}
\def\otsT{\mbox{$\bo_{m,k}^{(s)T}$}}
\def\oksT{\mbox{$\bo_m^{(s)T}$}}

\def\cSkrOne{\mbox{$S_{kr}^{(1)}$}}
\def\cSkTwo{\mbox{$S_k^{(2)}$}}
\def\cSkrTwo{\mbox{$S_{kr}^{(2)}$}}

\def\bsT{\mbox{$b^{(s)T}$}}
\def\brsT{\mbox{$b_r^{(s)T}$}}

\def\Dks{\mbox{$\Delta_k^{(s)}$}}
\def\DksT{\mbox{$\Delta_k^{(s)T}$}}

\def\Xs{\mbox{$\X^{(s)}$}}
\def\Zs{\mbox{$\Z^{(s)}$}}
\def\xhap{\mbox{$\hat{x}_{\alpha\rho}[n]$}}
\def\qapmn{\mbox{$q_{\alpha\rho}^{(m)}[n]$}}
\def\qapm{\mbox{$q_{\alpha\rho}^{(m)}$}}
\def\thetap{\mbox{$\theta_1$}}

\def\G{\mbox{$\mathcal{G}$}}
\def\Gn{\mbox{$\G_n$}}
\def\A{\mbox{$A$}}
\def\Aks{\mbox{$A^{(s)}_{k}$}}
\def\Akst{\mbox{$A^{(s)T}_{k}$}}
\def\Bks{\mbox{$B^{(s)}_{k}$}}
\def\Bkst{\mbox{$B^{(s)T}_{k}$}}
\def\Ak{\mbox{$A_{k}$}}
\def\Atk{\mbox{$A^{T}_{k}$}}
\def\xtst{\mbox{$x^{(s)T}_t$}}

\def\threeHalves{\mbox{$\frac{3}{2}$}}
\def\oneQuarter{\mbox{$\frac{1}{4}$}}

\long\def\gobbleup#1{}

\def\argmin{\operatornamewithlimits{argmin}}
\def\argmin{\mathop{\rm argmin}}

\def\argmax{\operatornamewithlimits{argmax}}
\def\argmax{\mathop{\rm argmax}}

\def\phih{\mbox{$\phi_{\mathbf{h}}$}}

\def\bE{\mbox{$\mathbf{E}$}}
\def\bq{\mbox{$\mathbf{q}$}}
\def\bt{\mbox{$\mathbf{t}$}}
\def\bbs{\mbox{$\mathbf{s}$}}
\def\bbbs{\mbox{$\mathbf{b}^{(s)}$}}
\def\bn{\mbox{$\mathbf{n}$}}
\def\bgs{\mathbf{g}_{1:K}^{(s)}}

\def\removal{\mbox{$\text{$\epsilon$--removal}$}}
\def\push{\mbox{$\text{push}$}}

\def\bO{\mbox{$\mathbf{O}$}}
\def\cK{\mbox{$\mathbb{K}$}}
\def\cR{\mbox{$\mathbb{R}$}}

\def\cn{\mbox{$c[n]$}}
\def\ctn{\mbox{$\tilde{c}[n]$}}
\def\chn{\mbox{$\hat{c}[n]$}}
\def\xhn{\mbox{$\hat{x}[n]$}}
\def\xh{\mbox{$\hat{x}$}}
\def\Cz{\mbox{$C(z)$}}
\def\Ctz{\mbox{$\tilde{C}(z)$}}
\def\Chz{\mbox{$\hat{C}(z)$}}
\def\Ch{\mbox{$\hat{C}$}}
\def\ahi{\mbox{$\hat{a}_i$}}
\def\bhi{\mbox{$\hat{b}_i$}}
\def\cchi{\mbox{$\hat{c}_i$}}
\def\dhi{\mbox{$\hat{d}_i$}}
\def\ai{\mbox{$a_i$}}
\def\bj{\mbox{$b_j$}}
\def\bji{\mbox{$b^{-1}_i$}}
\def\bi{\mbox{$b_i$}}
\def\ci{\mbox{$c_i$}}
\def\di{\mbox{$d_i$}}
\def\ati{\mbox{$\tilde{a}_i$}}
\def\btj{\mbox{$\tilde{b}_j$}}
\def\bhj{\mbox{$\hat{b}_j$}}
\def\bhji{\mbox{$\hat{b}^{-1}_i$}}
\def\Qz{\mbox{$Q(z)$}}
\def\lra{\mbox{$\leftrightarrow$}}
\def\calL{\mbox{$\mathcal{L}$}}
\def\K{\mbox{$\mathcal{K}$}}
\def\calE{\mbox{$\mathcal{E}$}}
\def\xn{\mbox{$x[n]$}}
\def\xmin{\mbox{$x_{\text{min}}[n]$}}
\def\calG{\mbox{$\mathcal{G}$}}

\def\Qmz{\mbox{$Q^m(z)$}}
\def\Qm{\mbox{$Q^m$}}
\def\qmn{\mbox{$q^{(m)}[n]$}}
\def\qm{\mbox{$q^{(m)}$}}
\def\sumk{\mbox{$\disp\sum_k$}}
\def\suml{\mbox{$\disp\sum_{l=0}^{L-1}$}}
\def\Fz{\mbox{$F(z)$}}
\def\Fmz{\mbox{$F^m(z)$}}
\def\Fm{\mbox{$F^m$}}
\def\ointC{\mbox{$\disp\oint_{C}$}}
\def\Sonez{\mbox{$S_1(z)$}}
\def\sonen{\mbox{$s_1[n]$}}
\def\popt{\mbox{$\hat{p}$}}

\def\betas{\mbox{$\beta^*$}}
\def\zo{\mbox{$z_0$}}
\def\zhn{\mbox{$\hat{z}_n$}}
\def\zho{\mbox{$\hat{z}_0$}}
\def\Fhz{\mbox{$\hat{F}(z)$}}
\def\Fh{\mbox{$\hat{F}$}}
\def\Fhpz{\mbox{$\hat{F}'(z)$}}
\def\Fhp{\mbox{$\hat{F}'$}}
\def\gammas{\mbox{$\gamma^*$}}

\def\complex{\mathbb{C}}
\def\compstar{\mbox{$\complex^*$}}

\def\Hz{\mbox{$H(z)$}}
\def\Xz{\mbox{$X(z)$}}
\def\Xhz{\mbox{$\hat{X}(z)$}}
\def\Xh{\mbox{$\hat{X}$}}
\def\Htz{\mbox{$\tilde{H}(z)$}}
\def\Hhz{\mbox{$\hat{H}(z)$}}
\def\Hh{\mbox{$\hat{H}$}}
\def\Ghz{\mbox{$\hat{G}(z)$}}
\def\Gh{\mbox{$\hat{G}$}}
\def\fm{\mbox{$f^{(m)}$}}

\def\bwai{\mbox{$\bw_{\text{a},i}$}}
\def\bwcai{\mbox{$\bw^*_{\text{a},i}$}}
\def\bwaone{\mbox{$\bw_{\text{a},1}$}}
\def\bwatwo{\mbox{$\bw_{\text{a},2}$}}
\def\bwcaone{\mbox{$\bw^*_{\text{a},1}$}}
\def\bwcatwo{\mbox{$\bw^*_{\text{a},2}$}}
\def\bwqi{\mbox{$\bw_{\text{q},i}$}}
\def\bwqone{\mbox{$\bw_{\text{q},1}$}}
\def\bwqtwo{\mbox{$\bw_{\text{q},2}$}}

\def\fs{\mbox{$f_{\text{s}}$}}

\def\bTheta{\mbox{$\boldsymbol{\Theta}$}}

% Symbols for maximum negentropy beamforming.
\def\cJ{\mbox{$\mathcal{J}$}}
\def\bwa{\mbox{$\bw_{\text{a}}$}}

\def\lambdas{\mbox{$\lambda_{\text{s}}$}}
\def\fs{\mbox{$f_{\text{s}}$}}
\def\bks{\mbox{$\bk_{\text{s}}$}}
\def\sigw{\mbox{$\sigma_{\text{w}}^2$}}
\def\sigI{\mbox{$\sigma^2_{\text{I}}$}}
\def\rhosone{\mbox{$\rho_{\text{s}1}$}}
\def\bBc{\mbox{$\bB_{\text{c}}$}}
\def\Bc{\mbox{$B_{\text{c}}$}}

\def\expect{\cE}

\nc{\bwq}{\mbox{$\bw_{\text{q}}$}}

\def\conjYn{\mbox{$\bar{Y}_n^m(\Omega)$}}
\def\conjY0{\mbox{$\left[Y_n^m(\bOmega_0)\right]^*$}}
\def\Y0{\mbox{$Y_n^m(\bOmega_0)$}}
\def\conjYs{\mbox{$\left[Y_n^m(\bOmega_s)\right]^*$}}
\def\Ys{\mbox{$Y_n^m(\bOmega_s)$}}

\def\bVs{\mbox{$\mathbf{V}_{\text{sp}}$}}
\def\bVsH{\mbox{$\mathbf{V}_{\text{sp}}^H$}}
\def\bNs{\mbox{$\mathbf{N}_{\text{sp}}$}}
\def\bNsH{\mbox{$\mathbf{N}_{\text{sp}}^H$}}
\def\bYs{\mbox{$\mathbf{Y}_{\text{sp}}$}}
\def\bws{\mbox{$\mathbf{w}_{\text{sp}}$}}
\def\bwsH{\mbox{$\mathbf{w}_{\text{sp}}^H$}}
\def\bwsMVDR{\mbox{$\mathbf{w}_{\text{sp,MVDR}}$}}
\def\bwsMVDRH{\mbox{$\mathbf{w}_{\text{sp,MVDR}}^H$}}
\def\bXs{\mbox{$\mathbf{X}_{\text{sp}}$}}
\def\bXsH{\mbox{$\mathbf{X}_{\text{sp}}^H$}}
\def\bGamma{\mbox{$\mathbf{\Gamma}$}}

\def\SigsX{\mbox{$\Sigma_{\text{sp},\mathbf{X}}$}}
\def\SigsV{\mbox{$\Sigma_{\text{sp},\mathbf{V}}$}}
\def\SigsN{\mbox{$\Sigma_{\text{sp},\mathbf{N}}$}}
\def\iSigsX{\mbox{$\Sigma_{\text{sp},\mathbf{X}}^{-1}$}}
\def\iSigsV{\mbox{$\Sigma_{\text{sp},\mathbf{V}}^{-1}$}}
\def\iSigsN{\mbox{$\Sigma_{\text{sp},\mathbf{N}}^{-1}$}}
\def\sigs0{\mbox{$\sigma_{\text{sp},0}^2$}}
\def\bQiso{\mbox{$\mathbf{Q}_{\text{iso}}$}}

% \nc{\bwqi}{\mbox{$\bw_{\text{q},i}$}}
% \nc{\bwqone}{\mbox{$\bw_{\text{q},1}$}}
% \nc{\bwqtwo}{\mbox{$\bw_{\text{q},2}$}}

\maketitle
\begin{abstract}
The sparsely-gated Mixture of Experts (MoE) can magnify a network capacity with a little computational complexity. In this work, we investigate how multi-lingual Automatic Speech Recognition (ASR) networks can be scaled up with a simple routing algorithm in order to achieve better accuracy. More specifically, we apply the sparsely-gated MoE technique to two types of networks: Sequence-to-Sequence Transformer (S2S-T) and Transformer Transducer (T-T). We demonstrate through a set of ASR experiments on multiple language data that the MoE networks can reduce the relative word error rates by 16.3\% and 4.6\% with the S2S-T and T-T, respectively. Moreover, we thoroughly investigate the effect of the MoE on the T-T architecture in various conditions: streaming mode, non-streaming mode, the use of language ID and the label decoder with the MoE.   
\end{abstract}
\noindent\textbf{Index Terms}: Multi-lingual speech recognition, End-to-end model, Mixture of experts, Transformers

\section{Introduction}

\label{sec:introduction}
% introduce the topic in this paper,  multi-lingual ASR
There has been a great deal of attention to multi-lingual automatic speech recognition (ASR) that can recognize multiple languages~\cite{Schultz2001,Ghoshal2013,cui2015multilingual,chen2015multitask,watanabe2017language,toshniwal2018multilingual,waters2019leveraging,Kim2018,kannan2019large,wang2021unispeech,wang2021unispeech2,lahiri2021multilingual,Guar2021,zhou2021configurable,Karimi2022,li2021scaling,xiao2021scaling,bai2021joint,AConneau2020,Raghavan2021}. Recent advances of end-to-end (E2E) ASR techniques have simplified multi-lingual ASR with a single network~\cite{watanabe2017language,toshniwal2018multilingual,waters2019leveraging,lahiri2021multilingual,Guar2021,zhou2021configurable,xiao2021scaling,bai2021joint}. Such a multi-lingual ASR model has a potential of improving accuracy on low resource language and code-switching data~\cite{kannan2019large,Guar2021}. The majority of multi-lingual ASR work focuses on the efficient use of multi-lingual training data through self-supervised learning~\cite{lahiri2021multilingual,Karimi2022,AConneau2020}, combination of unsupervised and supervised learning~\cite{wang2021unispeech,wang2021unispeech2,bai2021joint,Raghavan2021}, meta-transfer learning~\cite{lahiri2021multilingual}, multi-task learning~\cite{cui2015multilingual,chen2015multitask,Kim2018}, adapter networks~\cite{kannan2019large,Hou2021} or configurable network~\cite{zhou2021configurable}.
% mention about a motivation of this work:
Nowadays a large amount of multiple language data is usually available for multi-lingual ASR. Yet, the capacity of conventional E2E ASR networks will not be sufficient to digest large-scale multi-lingual information including many domains~\cite{li2021scaling,xiao2021scaling}. In \cite{xiao2021scaling}, Xiao et al. built a gigantic dense network with a massive amount of data. Such a dense model is proven to be powerful but results in an explosion in training and inference costs. Recently, sparsely gated mixture of experts (MoE) networks have been proposed as a computationally efficient way to increase the model capacity~\cite{Shazeer2017,fedus2021switch}. In this work, we investigate the effect of the MoE on multi-lingual speech recognition accuracy. 

% Prior work in applying MoE to ASR
In~\cite{you2021speechmoe,you2021speechmoe2}, You et al. utilized a variant of MoEs to the mere stacks of the Transformer layers in the Connectionist Temporal Classification (CTC) framework for ASR. They reported a prominent improvement by modifying the loss terms and routing algorithm for mono-lingual ASR tasks. Performance on the multi-lingual data would be limited because of the simple Transformer architecture trained based on the CTC loss. For multi-lingual ASR, Gaur et al. investigated the effect of the MoE with the RNN-T network~\cite{Guar2021}. They proposed the mixture of informed experts (MIE) by specializing the experts on data from a particular language. Despite of a relatively small LSTM-based model, they achieved an improvement in word error rate (WER) of up to 9\% relative on lower-resource languages while maintaining performance on high resource languages. It is worth noting that~\cite{zhou2021configurable} attempts at minimizing the multi-lingual E2E network size by manually hardcoding assignment between a specific expert and language.

% Contribution of this work
In this work, we apply the MoE to two types of popular E2E Transformers, sequence-to-sequence Transformer (S2S-T) and Transformer Transducer (T-T), for multi-lingual ASR. Moreover, we further increase the capacity of those E2E Transformer networks in contrast to the small RNN-T~\cite{Guar2021} or fixed routing mechanism based on language input~\cite{zhou2021configurable}. We investigate the effect of recognition accuracy with respect to the network capacity increased with the MoE technique.  

In section~\ref{sec:MoE}, we review the Switch Transformer architecture that will be adopted in this work. Section~\ref{sec:e2e} describes the end-to-end ASR networks with the MoE for our studies. Section~\ref{sec:ex} describes ASR experiments on multiple languages. We conclude this work and describe our future plan in section~\ref{sec:conclusion}. 

\section{Sparsely-gated mixture of experts (MoE)}
\label{sec:MoE}
\begin{figure}[t]
  \vspace{-1.0em}
  \centering
  \includegraphics[width=\linewidth]{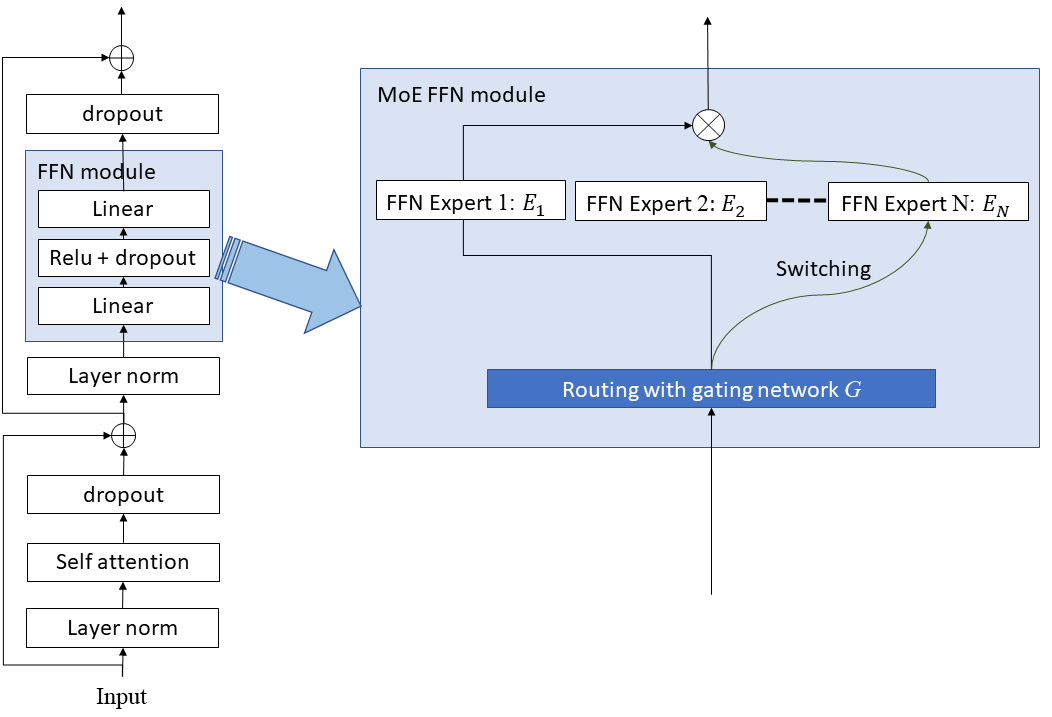}
  \caption{Schematic diagram of MoE Transformer encoder}
  \vspace{-1.8em}
  \label{fig:moe_block_chart}
\end{figure}
As the name indicates, the MoE layer typically consists of a set of $N$ \emph{expert networks} and routes an input representation $\bx$ to the $K$-best experts out of $N$ experts based on the \emph{gating network} output~\cite{Shazeer2017}. Denoting $G(\bx)_i$ and $E_i(\bx)$ as the $i$-th best output of the gating network and the output of the $i$-th best expert network on input $\bx$, the MoE layer output $\by$ can be expressed in the weighted sum of the experts as
\begin{equation}
\by = \sum_{i}^{K} G(\bx)_i E_i(\bx)
\vspace{-1.1em}
\label{eq:moe}
\end{equation}
Depending on the sparsity of $G(\bx)$, computation of the MoE can be saved. 

In contrast to~\cite{Shazeer2017}, Fedus et al.~\cite{fedus2021switch} adopted a simple switching strategy, routing the data to only a single expert instead of using the multiple experts. They applied a simple switching router to the Feed-Forwarding Network (FFN) layer. Figure~\ref{fig:moe_block_chart} shows a block chart of the MoE with a switching router, also known as Switch Transformer. Since only the single expert is selected, Switch Transformer does not increase computational complexity much. An only additional cost will be computing the gating function.   

% Capacity factor
A set of experts or each expert can be assigned to each device and feed-forwarding computation is performed on the selected expert. The data samples are routed to the expert with the highest router probability. Here, we follow~\cite{ fedus2021switch} to control the number of the data samples computed with each expert through the \emph{capacity factor} as 
\begin{eqnarray*}
\text{expert capacity} = \left(\frac{\text{samples per batch}}{\text{number of experts}} \right)  \times \text{capacity factor}
\vspace{-1.1em}
\label{eq:expert_capacity}
\end{eqnarray*}
When the capacity factor is greater than 1.0, buffer will be created to handle the case that the input data are not perfectly balanced across experts. 
% small capacity
If the input data samples are unevenly distributed to the expert, some of data samples will not be assigned to any experts and data overflows occur. The overflown data will be passed directly to the next layer through the residual connection without being processed by any experts. In other words, the data will be dropped at the expert layer in the case that too many samples are routed to a single expert.
A larger capacity factor keeps the data overflow from happening, but increasing the capacity factor will lead to the inefficient use of computation and memory. In the following experiment, we set the capacity factor to 1.5.  

To balance a data load across the experts, the auxiliary loss is typically added~\cite{ Shazeer2017,fedus2021switch}. Following~\cite{ fedus2021switch}, we compute the auxiliary loss given a batch $\mathcal{B}$ and $T$ samples as:
\begin{eqnarray}
\text{aux. loss} =  \alpha N \cdot \sum_{i=1}^{N} f_i \cdot P_i
\vspace{-1.1em}
\label{eq:loss1}
\end{eqnarray}
where $f_i$ is the fraction of samples dispatched to expert $i$,
\begin{eqnarray}
f_i =  \frac{1}{T} \sum_{\bx \in \mathcal{B}} \mathbb{1}\{ \argmax G(\bx), i \}
\vspace{-1.1em}
\label{eq:loss2}
\end{eqnarray}
$P_i$ indicates the fraction of the router probability allocated for expert $i$ and is computed by averaging the probability of routing sample $\bx$ to expert $i$ as,
\begin{eqnarray}
P_i =  \frac{1}{T}  \sum_{\bx \in \mathcal{B}} G(\bx)_i
\vspace{-1.1em}
\label{eq:loss3}
\end{eqnarray}
In this work, we set $\alpha=\text{0.01}$. 

We also introduce the \emph{switching jitters}, a small variable to control the multiplicative fraction to the gate input as $\bx *= \text{uniform}(1-\epsilon, 1+\epsilon)$. We use $\epsilon=\text{0.01}$ in all the experiments reported in the next section. 

%
%{\color{red}  TODO: describe some implementation details @Robert}
%

\section{End-to-end speech recognition network}
\label{sec:e2e}
In this section, we first describe the S2S Transformer network and explain how the MoE is applied to the S2S network. We then describe the T-T network with the MoE. 

\subsection{Sequence-to-sequence Transformer (S2S-T)}
\label{subsec:s2s}
\begin{figure}[t]
  \vspace{-1.0em}
  \centering
  \includegraphics[width=0.9\linewidth]{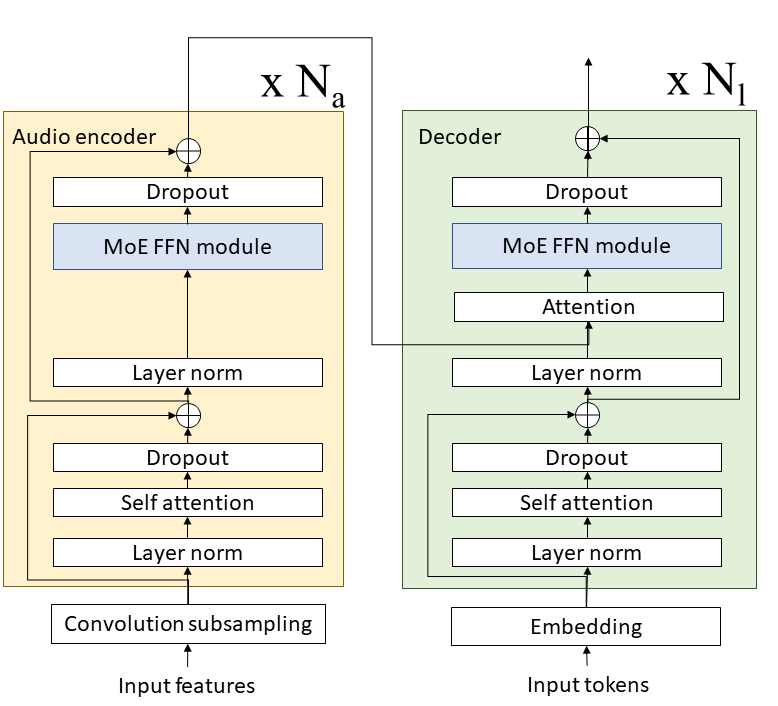}
  \caption{MoE Sequence-to-sequence Transformer (S2-T)}
  \label{fig:s2s}
  \vspace{-1.5em}
\end{figure}

Figure~\ref{fig:s2s} depicts the network architecture of the S2S-T for end-to-end ASR. As illustrated in Figure~\ref{fig:s2s}, the S2S-T consists of the convolution subsampling layer, the audio encoder, the label token embedding layer and the decoder. 
 Each encoder block is composed of two sub-layers: the self-attention module followed by the MoE FFN module. We employ Multi-Headed Attention (MHA) used in~\cite{li2020comparison} here. Each decoder layer consists of three sub-layers: self-attention, encoder-decoder attention and the MoE FFN module. As shown in Figure~\ref{fig:s2s}, each Transformer layer has residual connections around each of the sub-layers, layer normalization and dropout. Note that we employ the pre-layer norm residual units, passing the data before layer normalization. The S2S-T network used here does not mask attention input so that the attention module can observe the whole context. %For a streaming model study, we use T-T explained in the next section. 

The FFN module is conventionally composed of the linear layer, a nonlinear activation function followed by the linear layer without the MoE layer, as shown on the left side in Figure~\ref{fig:moe_block_chart}. In contrast to normal practice, we introduce the MoE FFN module, as depicted in the right block in Figure~\ref{fig:moe_block_chart}. We also apply ReLU activation and dropout between the MoE layers. 

\subsection{Transformer Transducer (T-T)}
\label{subsec:t-t}

The S2S-T can achieve better accuracy in many sequence-to-sequence classification tasks. However, maintaining accuracy with low latency in a streaming task is still a challenge~\cite{inaguma2020minimum}. On the contrary, the T-T network is suitable for the streaming task~\cite{QZhang2020}. For that reason, we also investigate the effect of the MoE on the T-T network. 

The main difference between the T-T and S2S network is that the transducer network provides a probability distribution over the label space at each time step while the S2S model attends over the whole input for every prediction in the output label sequence. This is why the use of the T-T is much more straightforward than the adoption of the S2S network in the streaming tasks. The output label of the transducer network includes an additional null label to indicate the lack of output at that time step in a similar manner with the CTC framework~\cite{graves2006ctc}. In contrast to CTC, the transducer networks can condition the label distribution on the label history~\cite{graves2012sequence}.  

\begin{figure}[t]
  \vspace{-1.0em}
  \centering
  \includegraphics[width=\linewidth]{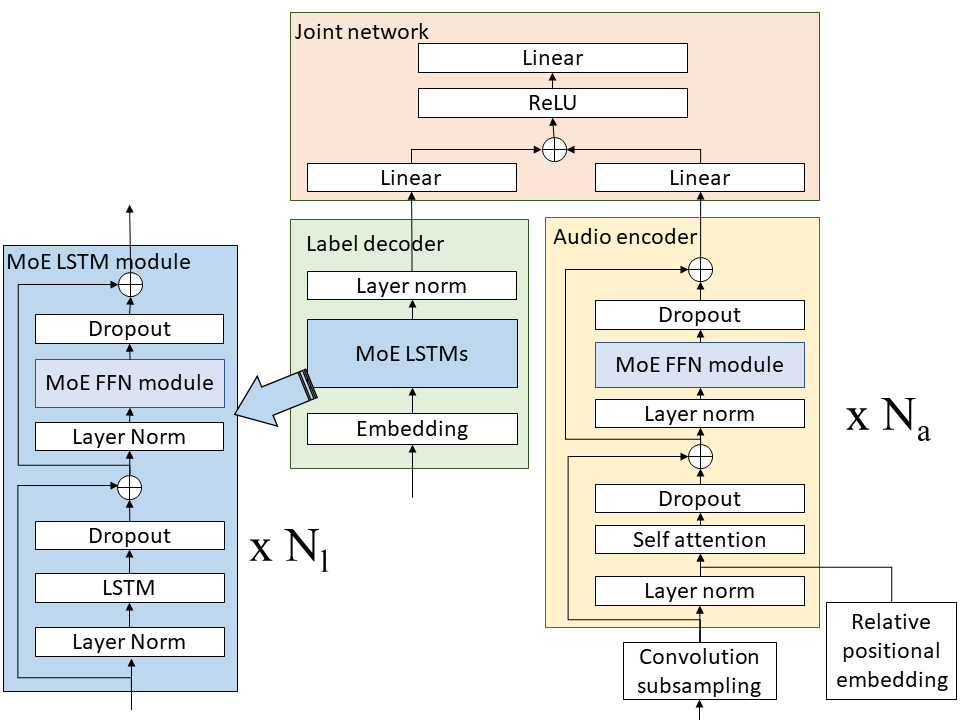}
  \caption{MoE Transformer Transducer (T-T)}
  \vspace{-1.5em}
  \label{fig:tt}
\end{figure}

Figure~\ref{fig:tt} shows the T-T architecture adopted in this work. 
As illustrated in Figure~\ref{fig:tt}, our T-T network consists of the audio encoder, label decoder and joint network. The audio encoder block is composed of the convolution subsampling layer followed by a stack of the Transformer blocks with the relative positional embedding~\cite{li2020comparison}. Each encoder layer contains two sub-layers: the MHA layer and the feed-forward network layer. Layer normalization is applied prior to the MHA layer. We use the relative position differences between input elements for multi-headed attention. In contrast to absolute position representations, the relative position representation is invariant to the total sequence length. This attention mechanism allows us to control the history length and latency of recognition output by masking the attention score of the left and right frames. To encourage gradient flow, we put a residual connection between the layer normalization input and dropout output for each sub-layer. In this work, we replace the conventional FFN layer with the MoE FFN layer. Our label decoder is composed of the embedding layer followed by a stack of LSTM layers~\cite{sak14_interspeech}. In the following experiment with T-T, we also consider replacing the LSTM projection layer in the label decoder with the MoE FFN module. Similar to the Transformer layer, we perform pre-layer normalization and post-dropout for the LSTM layer and the MoE FFN module. Residual connections are added to both layers to propagate gradient information. The joint network linearly combines the audio encoder output at every time step and the label decoder output given the previous non-blank output label sequence. The combined output is fed to the ReLU activation function, a linear layer and the Softmax layer. The transducer loss is efficiently computed by using the forward algorithm~\cite{graves2012sequence}. 
%\begin{figure}[t]
%  \centering
%  \includegraphics[width=0.5\linewidth]{figs/lstm.png}
%  \caption{MoE LSTM used for the label decoder}
%  \vspace{-0.8em}
%  \label{fig:moe-lstm-label}
%\end{figure}

\section{Experiment}
\label{sec:ex}
\subsection{Datasets}

The dataset used for training consists of 10 languages totaling approximately 75 thousand hours. As shown in Table~\ref{tab:traindata}, our training dataset is comprised of data from English, Spanish, Romanian, German, Italian, French, Portuguese, Dutch, Polish and Greek. It is worth noting here that we report the experimental results for multiple variants of English (US and UK), Spanish (ES-ES and ES-MX) and Romanian (RO). Table~\ref{tab:testdata} tabulates the statistics of our test set for each locale: number of utterances and number of words. The dataset of each locale also contains not only various speakers but also different speech recognition tasks such as command-and-control tasks in mobile, office and car scenarios, Cortana phrases, dictation, conversational speech in telecommunication and so on. We consider the English test, Spanish test and Romanian test as a high-resource task, mid-resource task and low-resource task, respectively. 

For training a network, we further split the whole training data set into training and validation sets in order to determine the convergence. The amount of training data for each language is different as shown in Table \ref{tab:traindata}. We, thus, sample the lower resource data more frequently to balance the language data distribution during training. 
For all the experiments reported here, we used a 80-dimensional log filter-bank energy feature extracted at an interval of every 10\textit{ms}. 10014 unique BPE~\cite{sennrich2016} tokens are used for covering vocabulary of 10 languages. 
\begin{table}[t]
\centering
{\footnotesize
\captionsetup{justification=centering}
\begin{tabular}{cc}
\hline
\hline
 Language & Hrs \\
\hline
\hline
English (EN) & 38585  \\
\hline
Spanish (ES) & 7584  \\
\hline
German (DE) &  4893   \\
\hline 
French (FR) &  5955 \\
\hline
Italian (IT) & 6580   \\
\hline
Portuguese (PT) & 4302  \\
\hline
Polish (PL) &  2112 \\
\hline
Greek (EL) & 2190 \\
\hline
Romanian (RO) &  1899 \\
\hline
Dutch (NL) &  880 \\
\hline
\hline
\end{tabular}
%\vspace{-0.8em}
}
\caption{Duration of Training data: 10 language data}
\vspace{-1.8em}
\label{tab:traindata}
\end{table}

\begin{table}[t]
\centering
\captionsetup{justification=centering,margin=1cm}
%\resizebox{0.45\textwidth}{!}{
{\footnotesize
\begin{tabular}{cc cc}
\hline
\hline
 Language & Code & No. utterances & No. words \\
\hline
\hline
\multirow{2}*{English} & EN-US & 219965 & 1374748 \\
\cline{2-4}
 & EN-UK &  16743 & 79191 \\
\hline
\multirow{2}*{Spanish} & ES-MX & 15279 & 124742 \\
\cline{2-4}
 & ES-ES & 19275 & 150980  \\
\hline
Romanian & RO & 16626	& 365828 \\
\hline
\hline
\end{tabular}
}
%}
%\vspace{-0.8em}
\caption{Test dataset details}
\vspace{-1.8em}
\label{tab:testdata}
\end{table}

\subsection{Experiments}

\begin{table*}[t]
   %\begin{minipage}{\textwidth}
	\centering
	{\footnotesize
	\begin{tabular}{|c|c|c|c||c|c|c|c|c||c||l|}
	   \hline
	   Model & \#experts & \#params & Lang. ID & EN-US & EN-UK & RO & ES-ES & ES-MX & Overall & \\ \hline\hline
S2S Transformer & N/A & 98.9M & No &
10.7 & 9.44 & 23.1 & 15.0 & 15.3 & 13.43 & $_{1\text{st}}$\\
       \hline
       \multirow{3}*{MoE S2S Transformer} & 24 & 677M & No & 
9.68 & 9.31 & 18.9 & 14.1 & 14.7 & 11.92 & $_{2\text{nd}}$\\ 
       \cline{2-11}                             & 72 & 1.88B & No &
9.25 & 7.83 & 18.2 & 12.3 & 13.6 & 11.24 & $_{3\text{rd}}$\\ 
       \cline{2-11}                             & 120 & 3.09B & No &
9.00 & 7.83 & 18.5 & 13.6 & 13.6 & 11.29 & $_{4\text{th}}$\\
       \hline
	\end{tabular}
	}
	\caption{WERs of sequence-to-sequence Transformers (S2S-T) with respect to the number of experts.}
	\vspace{-1.5em}
    \label{tab:wer_s2s}
\end{table*}
\begin{table*}[t]
	\centering
	{\footnotesize
	\begin{tabular}{|c|c|c|c|c|c||c|c|c|c|c||c||l|}
	   \hline
	    \multirow{2}*{Model} & \multicolumn{2}{|c|}{\#exprts} &  \multirow{2}*{\#params} &  \multirow{2}*{Streaming} &  \multirow{2}*{Lang. ID} &  \multirow{2}*{EN-US} &  \multirow{2}*{EN-UK} &  \multirow{2}*{RO} &  \multirow{2}*{ES-ES} &  \multirow{2}*{ES-MX} &  \multirow{2}*{Overall} & \\ 
	  \cline{2-3} & Audio& Label & & & & & & & & & &\\ \hline\hline
       \multirow{2}*{T-T} & \multirow{2}*{N/A} & \multirow{2}*{N/A} & 87.3M & Yes & No & 
11.2 & 9.15 & 23.3 & 16.4 & 17.6 & 14.02 & $_{1\text{st}}$ \\
     \cline{4-13} & & & 87.7M & Yes & Yes & 
10.8 & 8.90 & 22.8 & 15.1 & 15.6 & 13.45 & $_{2\text{nd}}$\\
       \cline{4-13} & & & 87.7M & No & Yes & 
9.76 & 7.54 & 18.7 & 11.0 & 11.5 & 11.46 & $_{3\text{rd}}$\\ 
       \hline
       \multirow{2}*{MoE T-T} & \multirow{2}{*}{24} & \multirow{2}*{N/A} & \multirow{2}*{521M} & Yes & Yes & 
9.58 & 8.57 & 21.4 & 14.5 & 14.8 & 12.87 & $_{4\text{th}}$\\
       \cline{5-13} & & & & No & No & 
9.83 & 7.74 & 17.5 & 11.3 & 12.2 & 11.35 & $_{5\text{th}}$\\
       \cline{5-13} & & & & No & Yes & 
9.52 & 7.34 & 16.9 & 10.7 & 11.3 & 10.93 & $_{6\text{th}}$\\ 
       \cline{3-13} & & 24 & 621M & No & Yes & 9.51 & 7.38 & 16.8 & 10.5 & 11.1 & 10.93 & $_{7\text{th}}$ \\
       \hline
       Deeper T-T & N/A & N/A & 144M & Yes & Yes & 
10.0 & 7.86 & 20.4 & 13.0 & 13.7 & 12.18 & $_{8\text{th}}$\\
       \hline
       \multirow{2}*{\begin{minipage}{0.45in}Deeper MoE T-T\end{minipage}} & 24 & N/A & 1.01B & Yes & Yes & 
10.1 & 8.18 & 19.0 & 12.8 & 13.1 & 12.04 & $_{9\text{th}}$\\ 
       \cline{2-13} & 72 & N/A & 2.82B & Yes & Yes & 
9.67 & 7.88 & 19.3 & 12.5 & 13.3 & 11.79 & $_{10\text{th}}$\\
       \cline{2-13} & 24 & N/A & 1.01B & No & Yes & 
9.17 & 6.89 & 14.8 & 9.50 & 10.1 & 10.17 & $_{11\text{th}}$\\
       \cline{2-13} & 72 & N/A & 2.82B & No & Yes & 
9.19 & 6.99 & 14.8 & 9.42 & 10.1 & 10.17 & $_{12\text{th}}$\\
      \hline
	\end{tabular}
	}
	\caption{WERs with various Transformer Transducer (T-T) architectures.}
	\vspace{-2.5em}
    \label{tab:wer_tt}
\end{table*}

\subsubsection{S2S-T model}
\label{subsubsec:s2s}
 The S2S-T baseline model has 18 encoder layers and 6 decoder layers. The dropout probability is 0.1 and the hidden size is 2048 in the encoder and decoder. Each MHA layer has 8 heads. 
The number of FLOPs for the MoE S2S-T network is matched to that of the S2S-T baseline network. The MoE FFN is applied to every two Transformer layers. 
All the models were trained with the AdamW optimizer. The learning rate schedule and batch size were kept the same for all the experiments. Training was done by using 24 GPUs with three ND A100 v4 Azure Virtual Machines (VMs) or three NCv3-series VMs.  

Table~\ref{tab:wer_s2s} shows the word error rate (WER) on each locale and the overall WERs for each model architecture: the conventional S2S-T model and its MoE versions with 24 experts, 72 experts and 120 experts. Here, the overall WER is computed by averaging the WERs weighted with the number of the test words in each locale set. In Table~\ref{tab:wer_s2s}, the second column and the third column indicate the number of experts and the number of trainable parameters, respectively. Language ID information is not used for generating the numbers in Table~\ref{tab:wer_s2s}. It is clear from Table~\ref{tab:wer_s2s} that the accuracy can be improved by applying the MoE FFN layer to each Transformer layer. In these experiments, the S2S-T model with 72 experts provides the best accuracy although the difference between 72 and 120 experts is not that significant. The dropout probability of each MoE FFN is set to 0.1 in the case of 24 experts and 72 experts while it is set to 0.4 in the case of the 120 experts. The dropout of 0.1 for the 120 experts provided the worse result because of the overfitting issue. 

\subsubsection{T-T model}
The convolution sampling layer for the T-T is composed of three sub-layers: two convolutional neural network (CNN) layers with kernel size (3,3) and strides (2,2) and a linear layer with the output size of 512. The first CNN has a single input channel and 512 output channels, and the second CNN has 512 input channels and 512 output channels. 
Our baseline audio encoder is composed of 18 identical layers. Each layer contains two sub-layers: the 512-dimensional attention layer with 8 heads and the FFN module with the dimension size of 2048. In the case of the streaming mode, the attention layers only attend 18 left-contexts and 4 right-contexts. In the non-streaming mode, whole utterance frames are attended by the network. 
Our baseline label decoder is composed of a 320-dimensional embedding and two layers of the unidirectional LSTM with 1024 hidden units. The hidden size of the MoE FFN module for the LSTM layer is 1024. 
The joint network contains two linear layers to project each input to a 512-dimensional vector and the output layer to project to 10,015 tokens including the blank symbol. 

Table~\ref{tab:wer_tt} shows the WER obtained with different types of the T-T models on each locale and the overall WER weighted-averaged over all the locales. 
Here, we also investigate the effect of language ID information on the model. As the language ID feature, we use the one-hot vector in the same way as~\cite{lahiri2021multilingual}. It is clear from the 1st and 2nd rows in Table~\ref{tab:wer_tt} that the use of the language ID improves recognition accuracy for the non-MoE T-T. It is also clear from the 2nd and 3rd rows in Table~\ref{tab:wer_tt} that the non-streaming T-T model can reduce the overall WER by 14.7\% relative to the streaming model. By comparing the numbers in the 2nd row with those in the 4th row and the WERs in the 3rd row with those in the 6th row, we can see that recognition accuracy can be improved by applying the MoE to the audio encoder only. Observe that the use of input language ID information can also improve recognition accuracy in the case of the MoE network by contrasting the WERs in the 5th row to those in the 6th row. We can see from the WERs in the 6th and 7th rows that the MoE FFN module in the LSTM layer did not make a big difference.  
 To investigate the effect of the MoE model for a larger capacity, we further increased the number of the encoder layers from 18 to 36 (labeled as Deeper T-T in Table~\ref{tab:wer_tt}). From Table~\ref{tab:wer_tt}, it is clear that the overall WER can be reduced from 13.45\% to 12.18\% by increasing the number of the audio encoder layers even without the MoE layer. It is also clear that the use of the MoE can further improve recognition accuracy. In the case of the audio encoder with 18 Transformer layers, the MoE layer reduces the relative WER by 4.3\% and 4.6\% for streaming and non-streaming modes, respectively. Moreover, the MoE layer reduces the relative WER by 3.2\% for the deeper audio encoder network. However, the gain obtained with the MoE T-T is not as large as that with the MoE S2S-T. From the fact that the use of the MoE in the label decoder has no impact on recognition accuracy, we draw a conclusion that the T-T architecture does not combine audio and label information well. In fact, the S2S-T architecture can fuse audio and language information better by attending over the audio context vectors in each label decoder block and bypassing information to the end through the residual connections.

\section{Conclusion}
\label{sec:conclusion}
In this work, we have scaled up the capacity of the multi-lingual ASR network with sparsely gated MoE. The MoE method used in this work increases computational complexity only for gating calculation; an increase in computational complexity is trivial. We have also demonstrated through a set of ASR experiments on the multiple locales that the MoE can improve the recognition accuracy. The MoE gives a larger gain with S2S Transformer than that with the T-T network. We are going to show how the multi-lingual MoE model can be compressed with knowledge distillation in~\cite{JWong2019ts,kumatani2020seqkd}.

\section{Acknowledgements}

%The ISCA Board would like to thank the organizing committees of the past INTERSPEECH conferences for their help and for kindly providing the template files. \\
The authors would like to thank Long Wu, Rui Jiang, Liyang Lu, Peidong Wang, Tianyu Wu, Nayuki Kanda, Yao Qian, Jinyu Li, Saeed Maleki, Baihan Huang, Weixing Zhang, Jesse Benson, Tim Harris, Yuan Yu and Mengchen Liu for technical discussions. We also would like to thank Ed Lin, Michael Zeng, Xuedong Huang and Yuan Yu for their project support.
\newpage
\bibliographystyle{IEEEtran}
\bibliography{mybib}

\end{document}